\documentclass{article}

\usepackage{microtype}
\usepackage{graphicx}
\usepackage{subfigure}
\usepackage{booktabs} 

\usepackage{hyperref}

\usepackage[accepted]{icml2019}

\usepackage{amsmath,amsfonts,amsthm} 
\usepackage{algorithm}
\usepackage{algorithmic}
\usepackage[T1]{fontenc}
\usepackage{enumitem}
\usepackage{multirow}
\usepackage{url}

\newtheorem{theorem}{Theorem}

\icmltitlerunning{Rethinking the Usage of Batch Normalization and Dropout}

\begin{document}

\twocolumn[
\icmltitle{Rethinking the Usage of Batch Normalization and Dropout in the Training of Deep Neural Networks}

\icmlsetsymbol{equal}{*}

\begin{icmlauthorlist}
	\icmlauthor{Guangyong Chen}{equal,tencent}
	\icmlauthor{Pengfei Chen}{equal,cuhk,tencent}
	\icmlauthor{Yujun Shi}{nku}
	\icmlauthor{Chang-Yu Hsieh}{tencent}
	\icmlauthor{Benben Liao}{tencent}	
	\icmlauthor{Shengyu Zhang}{cuhk,tencent}	
\end{icmlauthorlist}

\icmlaffiliation{cuhk}{The Chinese University of Hong Kong}
\icmlaffiliation{tencent}{Tencent Technology}
\icmlaffiliation{nku}{Nankai University}

\icmlcorrespondingauthor{Benben Liao}{bliao@tencent.com}

\icmlkeywords{Machine Learning, ICML}

\vskip 0.3in
]



\printAffiliationsAndNotice{\icmlEqualContribution} 

\begin{abstract}
	In this work, we propose a novel technique to boost training efficiency of a neural network. Our work is based on an excellent idea that whitening the inputs of neural networks can achieve a fast convergence speed. Given the well-known fact that independent components must be whitened, we introduce a novel Independent-Component (IC) layer before each weight layer, whose inputs would be made more independent. However, determining independent components is a computationally intensive task. To overcome this challenge, we propose to implement an IC layer by combining two popular techniques, Batch Normalization and Dropout, in a new manner that we can rigorously prove that Dropout can quadratically reduce the mutual information and linearly reduce the correlation between any pair of neurons with respect to the dropout layer parameter $p$. As demonstrated experimentally, the IC layer consistently outperforms the baseline approaches with more stable training process, faster convergence speed and better convergence limit on CIFAR10/100 and ILSVRC2012 datasets. The implementation of our IC layer makes us rethink the common practices in the design of neural networks. For example, we should not place Batch Normalization before ReLU since the non-negative responses of ReLU will make the weight layer updated in a suboptimal way, and we can achieve better performance by combining Batch Normalization and Dropout together as an IC layer.
\end{abstract}

\section{Introduction}
Deep neural networks (DNNs) have been widely adopted in many artificial-intelligence systems due to their impressive performance. The state-of-the-art neural networks are often complex structures comprising hundreds of layers of neurons and millions of parameters.  Efficient training of a modern DNN is often complicated by the need of feeding such a behemoth with millions of data entries. Developing novel techniques to increase training efficiency of DNNs is a highly active research topics. In this work, we propose a novel training technique by combining two commonly used ones, Batch Normalization (BatchNorm) \cite{ioffe2015batch} and Dropout \cite{srivastava2014dropout}, for a purpose (making independent inputs to neural networks) that is not possibly achieved by either technique alone. This marriage of techniques endows a new perspective on how Dropout could be used for training DNNs and achieve the original goal of whitening inputs of every layer \cite{le1991eigenvalues,ioffe2015batch} that inspired the BatchNorm work (but did not succeed). 

Despite a seminal work \cite{le1991eigenvalues} advocating the benefits of whitening, this technique has largely been limited to preprocessing input data to a neural network. An attempt to implementing the whitening idea at every activation layer turned out to be computationally demanding and resulted in the proposal of BatchNorm instead. The method works by scaling the net activations of each activation layer to zero mean and unit variance for performance enhancement. The normalization procedure significantly smooths the optimization landscape in the parameter space \cite{Bjorck2018Understandbatch} to improve training performance. Due to BatchNorm's easy implementation and success, the attention was diverted away from the original goal of whitening.

The motivation of this work is to resume the pursuit of devising a computationally efficient approach to whitening inputs of every layer.  Given the well-known fact that the independent activations must be whitened, we attempt to make the net activations of each weight layer more independent. This viewpoint is supported by a recent neural scientific finding: representation power of a neural system increases linearly with the number of independent neurons in the population \cite{moreno2014information,beaulieu2018enhanced}. Thus, we are motivated to make the inputs into the weight layers more independent.  As discussed in a subsequent section, the independent activations indeed make the training process more stable. 

An intuitive solution to generate independent components is to introduce an additional layer, which performs the independent component analysis (ICA) \cite{oja2006independent} on the activations. A similar idea has been previously explored in \cite{huang2018decorrelated}, which adopts zero-phase component analysis (ZCA) \cite{bell1997edges} to whiten the net activations instead of making them independent. In particular, ZCA always serves as the first step for the ICA methods, which then rotates the whitened activations to obtain the independent ones. However, the computations of ZCA itself is quite expensive, since it requires to calculate the eigen-decomposition of a $d_j\times d_j$ matrix with $d_j$ being the number of neurons in the $j$-th layer. This problem is more serious for wide neural networks, where a large number of neurons often dwell in an intermediate layer.

While attacking this challenging problem, we find that BatchNorm and Dropout can be combined together to construct independent activations for neurons in each intermediate weight layer. To simplify our presentation, we denote the layers \emph{\{-BatchNorm-Dropout-\}} as an Independent Component (IC) layer in the rest of this paper.
Our IC layer disentangles each pair of neurons in a layer in a continuous fashion - neurons become more independent when our layer is applied (Section~\ref{mutual-info}). Our approach is much faster than the traditional whitening methods, such as PCA and ZCA as mentioned above. 
An intuitive explanation of our method is as follows.
The BatchNorm normalizes the net activations so that they have zero mean and unit variance just like the ZCA method. Dropout constructs independent activations by 
introducing independent random gates for neurons in a layer, 
which allows neurons outputing its value with probability $p$, or shuts them down by outputting zero otherwise. 
Intuitively, the output of an neuron conveys little information from other neurons.
Thus, we can imagine that these neurons become statistically independent from each other. As theoretically demonstrated in Section~\ref{mutual-info}, our IC layer can reduce the mutual information between the outputs of any two neurons by a factor of $p^2$ and reduce the correlation coefficient by $p$, where $p$ is the Dropout probability. To our knowledge, such a usage of Dropout has never been proposed before. 

\begin{figure}
	\begin{center}
		\includegraphics[width=0.9\linewidth]{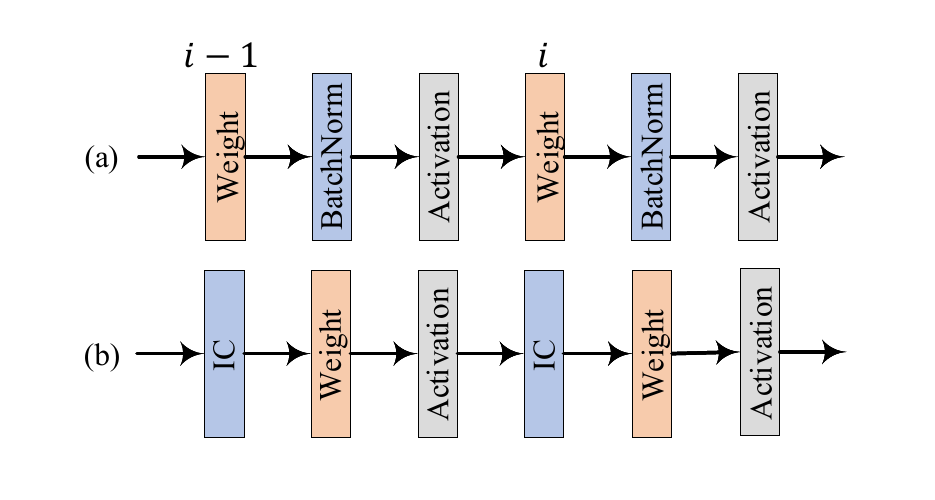}
	\end{center}
	\vspace{-8mm}
	\caption{(a) The common practice of performing the whitening operation, or named as batch normalization, between the weight layer and the activation layer. (b) Our proposal to place the IC layer right before the weight layer. }
	\label{fig_intro}
\end{figure}

Unlike the traditionally unsupervised learning methods, such as ICA and ZCA, we are not required to recover the original signals from the independent features or ensure the uniqueness of these features, but only required to distill some useful features, which can help fulfill the supervised learning tasks.
Our analysis, presented in Section \ref{Sec_Pre}, shows that
the proposed IC layer should be placed before the weight layer, instead of the activation layer as presented in \cite{ioffe2015batch,ioffe2017batch}. To evaluate the practical usages of our IC layer, we modify the famous ResNet architectures \cite{he2016deep,he2016identity} with our layer, and find that their performance can be further improved. As demonstrated empirically on the CIFAR10/100 and ILSVRC2012 datasets, the proposed IC layer can achieve more stable training process with faster convergence speed, and improve the generalization performance of the modern networks. 

The implementation of our IC layer makes us rethink the common practice of placing BatchNorm before the activation function in designing DNNs. The effectiveness of doing so for BatchNorm has been argued by some researchers, but no analysis has been presented to explain how to deal with the BatchNorm layer. The traditional usage of BatchNorm has been demonstrated to significantly smooth the optimization landscape, which induces a more predictive and stable behavior of the gradients \cite{santurkar2018does}. However, as discussed in Section \ref{Sec_Dis}, such usage of BatchNorm still forbids the network parameters from updating in the gradient direction, which is the fastest way for the loss to achieve minimum, and actually presents a zigzag optimization behavior.
Moreover, different from prior efforts \cite{li2018understanding} where BatchNorm and Dropout are suggested to be used simultaneously before the activation layer,
our analysis provides a unique perspective that BatchNorm and Dropout together play a similar role as the ICA method and should be placed before the weight layer, which admits a faster convergence speed when training the deep neural networks. Both theoretical analysis and experimental results indicate that BatchNorm and Dropout together should be combined as the IC layer, which can be widely utilized for training deep networks in future. 

To summarize, we list our main contributions as follows:
\begin{itemize}
	\item We combine two popular techniques, Batch Normalization and Dropout, in a newly proposed Independent-Component (IC) layer. We rigorously prove that the IC layer can reduce the mutual information and the correlation coefficient between any pair of neurons, which can lead to a fast convergence speed. 
	\item To corroborate our theoretical analysis, we conduct extensive experiments on CIFAR10/100 and ILSVRC2012 datasets. The results convincingly verify that our implementation improves the classification performance of modern networks in three ways: i) \textbf{more stable training process}, ii) \textbf{faster convergence speed}, and iii) \textbf{better convergence limit}.
	
\end{itemize}

\section{Preliminary: the Influence of Uncorrelated Components}
\label{Sec_Pre}
In this section, we  discuss the influence of uncorrelated inputs on the training of DNNs. 
Following \cite{le1991eigenvalues}, 
we consider a vanilla neural network consisting of a stack of linear layers. 
Let $A$ denote the parameter of the approximation function, whose inputs and outputs are $x$ and $y$ respectively. Recall the results presented in \cite{le1991eigenvalues}, we want to minimize the objective function 
\begin{equation}
	\min_{A}\sum_{i=1}^{n} \|y_i-Ax_i\|_2^2,
\end{equation}
then based on the gradient method, the optimal solution of $A$ must satisfy
\begin{equation}
	A\sum_{i=1}^{n}x_ix_i^T = \sum_{i=1}^{n}yx_i^T.
	\label{Eq_OptimalCon}
\end{equation}
It is obvious that if $x_i$ lies in the local subspace, $\sum_{i=1}^{n}x_ix_i^T $ can be represented by a low-rank matrix, which means multiple optimal solutions can be found for Eq. (\ref{Eq_OptimalCon}). Thus, the training process of DNNs may encounter some unstable cases. Moreover, as claimed in \cite{le1991eigenvalues}, when we update $A$ with the gradient method, the convergence speed depends on the ratio of largest eigenvalue to the smallest one of the Hessian matrix $\sum_{i=1}^{n}x_{i}x_i^T$. Thus, if each entry in $x_i$ is uncorrelated to each other, we would get faster convergence speed of training DNNs. ZCA has been previously explored to whiten the activations. However, the computations of ZCA itself is quite expensive, especially when whitening the activations of wide neural networks. Given the well-known fact that the independent activations must be whitened, we can whiten the activations by making them more independent, which can be effectively realized by our IC layer.

\section{Generating Independent Components: IC Layer}
\begin{figure}
	\begin{center}
	\includegraphics[width=\linewidth]{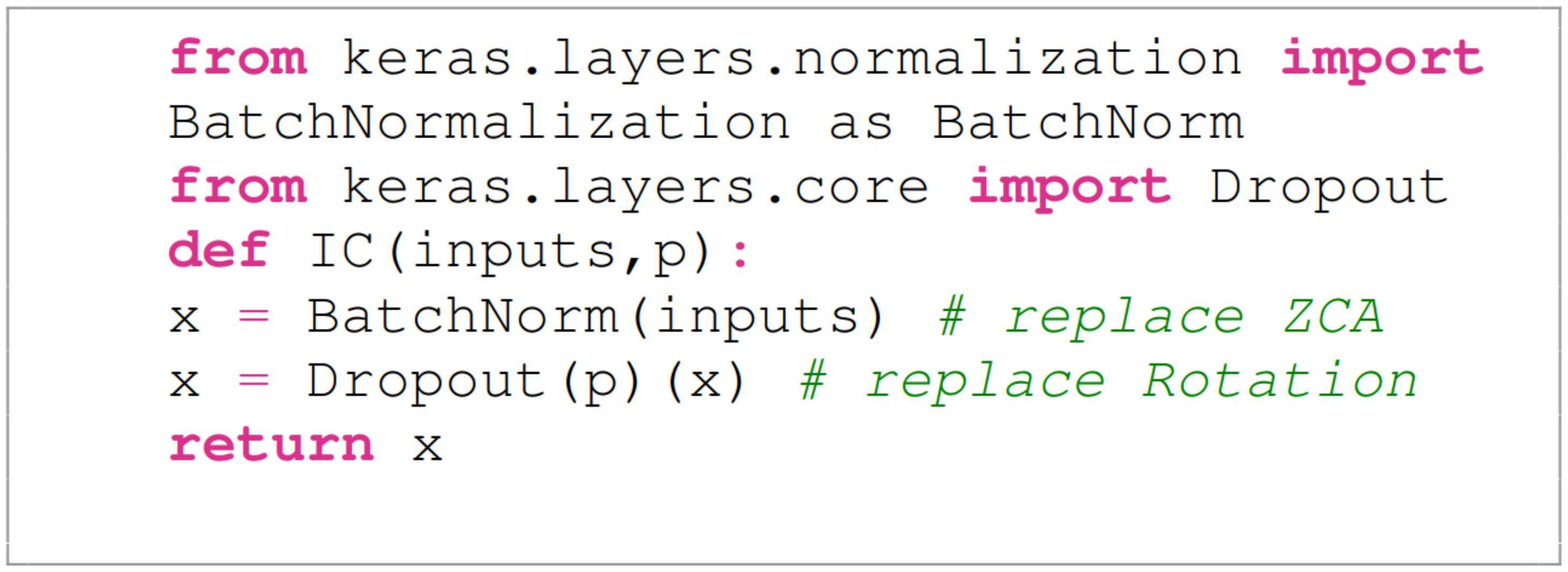}
	\end{center}
	\vspace{-2mm}
	\caption{Python code of the IC layer based on Keras}
	\label{fig_DropNorm_code}
\end{figure}

An intuitive idea to generate independent components is to implement the ICA method to net activations of layers. However, as discussed previously, the precise computation of the ICA method is too expensive. To address this issue, we propose an IC layer, which can be easily implemented by a few lines of python code based on Keras\footnote{https://keras.io/}, as shown in Fig.~\ref{fig_DropNorm_code}. Traditionally, the ICA can be implemented by two steps: the ZCA is implemented to decorrelate the input vector and the rotation operator is implemented to get the final independent components. Accordingly, we use BatchNorm to replace the ZCA and Dropout to replace the rotation step. 

In this section, we mainly focus on what roles Dropout plays in constructing a series of independent activations. Dropout is originally motivated by sampling a subnetwork from an exponential number of different "thinned" networks, which significantly reduces overfitting and gives major improvements over other regularization choices \cite{srivastava2014dropout}. In particular, Dropout introduces independent random gates for neurons in a layer, which allows neurons outputing its value with probability $p$, or shuts them down by outputting zero otherwise. Intuitively, the output of an neuron conveys little information from other neurons. In this section, we will theoretically depict the impact of Dropout on the training of DNNs by measuring the dependency among any two channels using mutual information and  correlation coefficient, resulting in a fast convergence speed. 

\subsection{Mutual Information and Entropy}\label{mutual-info}

It is well-known that the level of dependence between two
information sources $x,y$ can be quantified by their mutual information
$$I(x;y)=\sum_{x,y}P(x,y)\log\frac{P(x,y)}{P(x)P(y)}.$$
That is to say,
$x,y$ are independent if and only if $I(x;y)=0$.  By considering a random gate $g_i$ to independently modify the standardized activations $x_i$, which would remain the same  with the probability $p$ or set to be zero otherwise, we have the following theorem,
\begin{theorem}\label{thm}
	Let $g_i$ denote a family of independent random variables generated from the Bernoulli distribution 
	with its mean being $p$.  
	Let $\hat{x}_i = g_ix_i$.
	Then we have
	\begin{equation}
		\begin{split}
			&I(\hat{x}_i;\hat{x}_j) = p^2I(x_i,x_j), \forall i\neq j,\\
			&H(\hat x_i)=pH(x_i)+\epsilon_p,
		\end{split}
	\end{equation}
	where $H$ denotes the Shannon entropy and $\epsilon_p$ the entropy of the Bernoulli  distribution.
\end{theorem}

The theorem tells us that up to a small error $\epsilon_p$ (smaller than 'a bit'), the information contained in a neuron 
after dropout 
decays by a factor $p$.
In view of the decay factor $p^2$ for the mutual information between neurons, this loss of information is not significant. 
In the experimental results shown in later sections, we will balance the information loss and independence gain between neurons by choosing an optimal dropout parameter $p$.

We now give a proof of Theorem~\ref{thm}.

\begin{proof}
	For simplicity, we assume that the probabilities for which 
	variables $x_i,x_j$ being zeros are negligible.~\footnote{This always happens unless the distributions of $x_i,x_j$ are extremely chaotic.}
	We divide the summation 
	$I(\hat x_i;\hat x_j)$ into 3 parts: 
	$$I(\hat x_i;\hat x_j)=\sum_{a_1}+\sum_{a_2}+\sum_{a_1,a_2}.$$
	
	The first part concerns with computations when $\hat x_j$ are restricted to zero
	$$\sum_{a_1}=\sum_{a_1}P(\hat x_i=a_1,\hat x_j=0)\log\frac{P(\hat x_i=a_1,\hat x_j=0)}{P(\hat x_i=a_1)P(\hat x_j=0)},$$
	where $a_1$ runs over the range of the dropout input variable $x_i$.
	As the probability when $x_j$ being zero is negligible, we have 
	$P(\hat x_i=a_1,\hat x_j=0)=P(\hat x_i=a_1,g_j=0)$.
	The latter term equals to $P(\hat x_i=a_1)P(g_j=0)$
	since $g_j$ is generated independently from $x_i$ and $g_i$. 
	In summary, we have
	$$P(\hat x_i=a_1,\hat x_j=0)=P(\hat x_i=a_1,g_j=0)$$
	$$=P(\hat x_i=a_1)P(g_j=0)=P(\hat x_i=a_1)P(\hat x_j=0).$$
	Therefore, the two events $\hat x_i=a_1$ and $\hat x_j=0$ are independent. This implies that the first part $\sum_{a_1}$ is zero. 
	The same argument applies also to the second part $\sum_{a_2}=0$ which deals with the computations when $\hat x_i$ being zero instead.
	
	So the only contribution to the mutual information $I(\hat x_i;\hat x_j)$ is the third part
	$$\sum_{a_1,a_2}=\sum_{a_1,a_2}P(\hat x_i=a_1,\hat x_j=a_2)
	\log\frac{P(\hat x_i=a_1,\hat x_j=a_2)}{P(\hat x_i=a_1)P(\hat x_j=a_2)}.$$
	Since both $g_i,g_j$ are generated independently from $x_i,x_j$, we have
	$$P(\hat x_i=a_1,\hat x_j=a_2)=P(x_i=a_1,x_j=a_2,g_i=g_j=1)$$
	$$=p^2P(x_i=a_1,x_j=a_2).$$
	Similar computations apply to both $P(\hat x_i=a_1)$ and $P(\hat x_j=a_2)$
	$$P(\hat x_i=a_1)=pP(x_i=a_1),$$
	$$P(\hat x_j=a_2)=pP(x_j=a_2).$$
	As a result, this part is proportional to the mutual information $I(x_i,x_j)$ with factor $p^2$:
	$$\sum_{a_1,a_2}=p^2I(x_i;x_j).$$
	
	The overall computations show that by applying dropout to variables $x_i,x_j$, we succeed in decreasing their mutual information by a factor $p^2$:
	$$I(\hat x_i;\hat x_j)=p^2I(x_i;x_j).$$
	
	
	While mutual information has been decreased by a factor $p^2$, we do not lose too much information after dropout layer. 
	
	Indeed, consider the Shannon entropy of $\hat x_i$
	$$H(\hat x_i)=-\sum_a P(\hat x_i=a)\log P(\hat x_i=a)$$
	and divide the sum to first part where $a=0$ and the second part where $a\neq 0$
	$$H(\hat x_i)=-P(\hat x_i=0)\log P(\hat x_i=0)+\sum_{a\neq 0}.$$
	Since $P(\hat x_i=0)=P(g_i=0)=1-p,$
	the first part is $-(1-p)\log (1-p)$. For the second part $\sum_{a\neq 0}$, since 
	$$P(\hat x_i=a)=P(x_i=a,g_i=1)=pP(X_i=a)$$ for $a\neq 0$, it is equal to
	$$=-\sum_{a\neq 0}pP(x_i=a)\log pP(x_i=a)$$
	$$=-\sum_{a\neq 0}pP(x_i=a)\log p-p\sum_{a\neq 0}P(x_i=a)\log P(x_i=a)$$
	$$=-p\log p +pH(x_i).$$
	Therefore, the Shannon entropy of neuron $\hat x_i$
	$$H(\hat x_i)=pH(x_i)+\epsilon_p,$$
	where $\epsilon_p$ is the entropy of the Bernoulli $g_i$.
\end{proof}

\subsection{Correlation Coefficient and Expectation}
To help readers further understand the influence of Dropout on the optimization process, we further employ the correlation coefficient to depict the dependence relationship among the standardized activations, $x_i$ and $x_j$, given by the $i$-th and $j$-th neurons, as  follows,
\begin{equation}
	c_{ij} = \mathbb{E}(x_ix_j).
\end{equation}
By considering a random gate $g_i$ to independently modify the standardized activations $x_i$, which would remain the same  with the probability $p$ or set to be zero otherwise, we can reformulate the correlation as,
\begin{equation}
	\hat{c}_{ij} = \frac{1}{\sigma_{i}\sigma_{j}}\mathbb{E}(g_ix_ig_jx_j),
\end{equation}
where $\sigma_{i}=\sqrt{\mathbb{E}(g_{i}x_{i})^2}$ denotes the standard variance of $x_i$. Since $g_i$ is generated from a Bernoulli distribution, which is independent from the generation of $x_i$, then we have $\sigma_i^2=\mathbb{E}g_{i}^2x_{i}^2 = \mathbb{E}g_{i}^2\mathbb{E}x_{i}^2$. It is obvious that $g_{i}^2$ follows the same distribution as  the random gate $g_{i}$ with its expectation as $p$. Since $x_i$ has been standardized by  BatchNorm, we have $\mathbb{E}x_{i}^2= 1$ . Then we can get $\sigma_i^2=p$. Moreover, because $g_{i}$ is a random variable generated independent from each other and the samples $x_{i}$, then $g_{i}g_{j}$ is also independent from the random variable $x_{i}x_{j}$,  we can get
\begin{equation}
	\mathbb{E}[(g_{i}g_{j}-\nu_{ij})(x_{i}x_{j}-\mu_{ij})] = 0,
	\label{Eq_indep_gate_activation}
\end{equation}
where $\nu_{ij}$ and $\mu_{ij}$ are the expectations of the random variable $g_{i}g_{j}$ and $x_{i}x_{j}$, respectively. Since $g_{i}$ and $g_{j}$ are generated independently from each other, we have $\nu_{ij} = \mathbb{E}(g_{i}g_{j})=\mathbb{E}g_{i}\mathbb{E}g_{j} =p^2$. From the fact that both $x_{i}$ and $x_{j}$ have zero mean, we have $\mu_{ij} = 0$. Then we can reformulate Eq. \ref{Eq_indep_gate_activation} as,
\begin{equation}
	\mathbb{E}[(g_{i}g_{j}-p^2)x_{i}x_{j}]= 0
\end{equation}
which leads to the following equation,
\begin{equation}
	\hat{c}_{ij} = \frac{1}{\sigma_{i}\sigma_{j}}p^2\mathbb{E}(x_{i}x_{j}) = pc_{ij}.
\end{equation}
Thus, it shows that we can linearly reduce the correlation among activations given by any two units by introducing an independent gate for each individual neuron, which dramatically save lots of computational resources compared with ZCA or ICA methods. 

However, the side effect is that the expected activation also decrease. Take $x_{i}$ for example, its expected value would become $px_i$ if the random gate is open with the probability $p$. Thus, we get that the smaller $p$ make units work more independently, but loss more information which should be transferred through the network.

\section{Experiments and Results}

\subsection{Reformulating ResNet Architectures}
\begin{figure}
	\begin{center}
		\includegraphics[width=\linewidth]{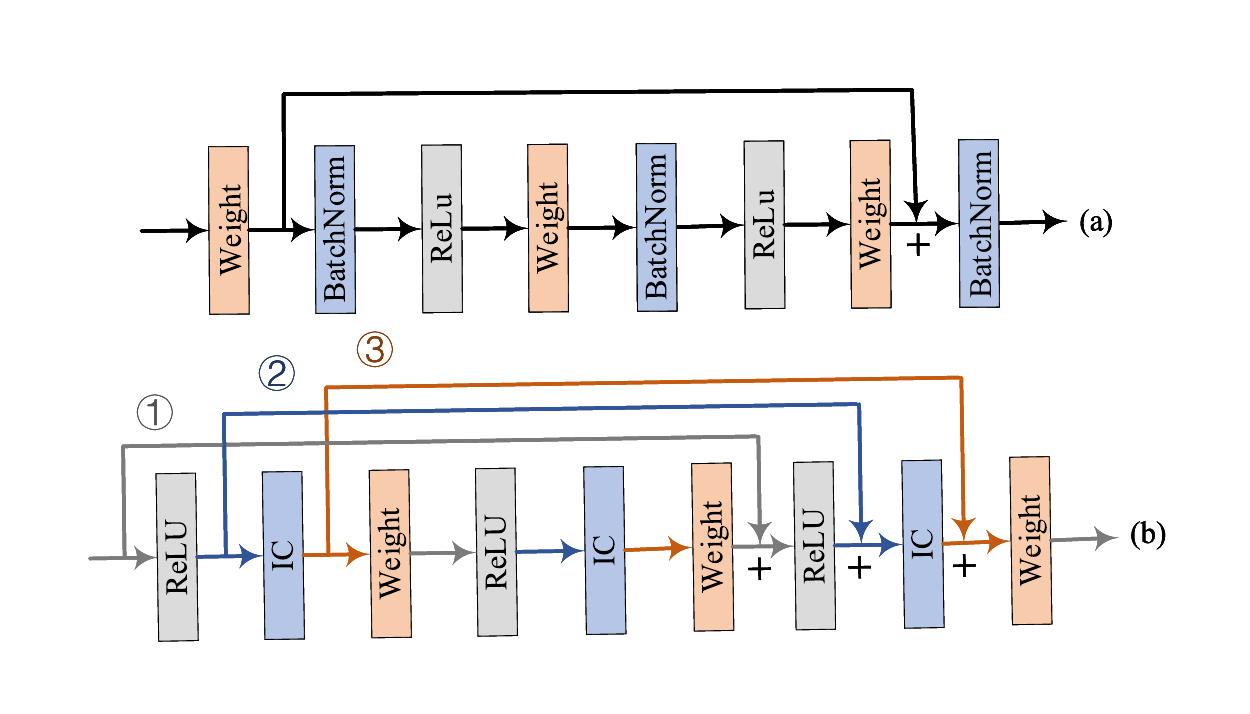}
	\end{center}
	\vspace{-8mm}
	\caption{(a) The classical ResNet architecture, where $'+'$ denotes summation. (b) Three proposed ResNet architectures reformulated with the IC layer.}
	\label{fig_ResNet_archtect}
\end{figure}
During recent years, many visual recognition tasks have greatly benefit from very deep models. However, a degradation problem arises when deep models start converging: with the network depth increasing, accuracy get saturated and then degrades rapidly. This problem may be caused by back-propagation gradient that can vanish or explode by the time it finally reaches the beginning of the deep network. To solve this issue, many efforts have been made by creating short paths from early layers to later layers. ResNet \cite{he2016deep} has attracted huge amount of attentions from the machine learning and computer vision community since it eases the training of extremely deep networks. Instead of making each few stacked layers directly fit a target mapping $f(x)$ as shown in Fig.~\ref{fig_intro}(a), ResNet explicitly lets these layers fit a residual mapping $f(x) - x$ with shortcut connections as shown in Fig.~\ref{fig_ResNet_archtect}(a) \cite{he2016identity}. It has been widely demonstrated that convolution networks can be substantially deeper by creating short paths from early layers to later ones.

\begin{table*}[t]
	\caption{The testing results of implementing ResNet and ResNet-B with the IC layer on the CIFAR10/100 datasets at the end of training.}
	\centering
	\begin{tabular}{ c|c|l|c|c }
		\hline
		Model& Depth&Layers in Residual Unit&CIFAR10&CIFAR100 \\ 
		\hline
		\multirow{8}{*}{ResNet} 
		& \multirow{4}{*}{110} 
		& \textcircled{1}. $2\times$\{ReLU-IC-Conv2D\} & 0.9361&0.725 \\
		& &\textcircled{2}. $2\times$\{IC-Conv2D-ReLU\} & 0.9352&0.7165 \\
		& & \textcircled{3}. $2\times$\{Conv2D-ReLU-IC\} & 0.9408&0.7174  \\ \cline{3-5}
		& & Baseline & 0.9292	&0.6893\\ \cline{2-5}
		& \multirow{4}{*}{164} 
		& \textcircled{1}. $2\times$\{ReLU-IC-Conv2D\} & 0.9395&0.7224 \\
		& &\textcircled{2}. $2\times$\{IC-Conv2D-ReLU\} & 0.9366&0.7273 \\
		& & \textcircled{3}. $2\times$\{Conv2D-ReLU-IC\} &  0.9411&0.7237  \\\cline{3-5}
		& & Baseline &  0.9339&0.6981  \\ \hline
		\multirow{8}{*}{ResNet-B} 
		& \multirow{4}{*}{110} &\textcircled{1}. $3\times$\{ReLU-IC-Conv2D\} & 0.9448&0.7563 \\
		&& \textcircled{2}. $3\times$\{IC-Conv2D-ReLU\} & 0.9425&0.7532 \\			
		& & \textcircled{3}. $3\times$\{Conv2D-ReLU-IC\} &  0.9433&0.7482  \\\cline{3-5}
		& & Baseline &  0.9333&0.7387  \\ \cline{2-5}
		& \multirow{4}{*}{164} &\textcircled{1}. $3\times$\{ReLU-IC-Conv2D\} & 0.9445&0.758 \\
		&& \textcircled{2}. $3\times $ \{IC-Conv2D-ReLU\} & 0.9424&0.7616 \\
		& & \textcircled{3}. $3\times$\{Conv2D-ReLU-IC\} &  0.9453&0.7548  \\\cline{3-5}
		& & Baseline &  0.9355&0.7465  \\ \hline
	\end{tabular}
	\label{tab_ResNet}
\end{table*}

In this paper, we attempt to implement the idea of ResNet with a stack of \emph{-IC-Weight-ReLU-} layers.  Following \cite{he2016identity}, we study three different types of residual units, each of which has a unique short path as shown in Fig.~\ref{fig_ResNet_archtect}(b), and aim to find the best residual unit. Due to the training cost of deep models, we further consider a bottleneck design \cite{he2016deep}, which replaces each residual unit by a bottleneck unit. For example, the corresponding bottleneck design of the first type of residual unit, consisting of $2\times $\emph{ReLU-IC-Weight} layers, is $3\times $\emph{ReLU-IC-Weight}, where the first and last weight layers are $1\times 1$ convolutions instead. 

we implement our modified ResNets architectures on the benchmark datasets, including CIFAR10/100 and ILSVRC2012 datasets, to evaluate the practical usages of the IC layer. Since our focus is on the behaviors of the IC layer in some modern architectures, but not pushing the state-of-the-art performance achieved on these benchmark datasets, we intentionally use the architectures recorded in the original papers \cite{he2016deep,he2016identity,li2018understanding} and follow the publicly published parameters configurations. For fair comparisons, we introduce a pair of trainable parameters for the IC layer, which scale and shift the values normalized by the BatchNorm, so that the modified ResNet will have the same amount of learnable parameters as the corresponding baseline architecture. 

\subsection{CIFAR10/100}
Both CIFAR datasets \cite{krizhevsky2009learning} consist of colorful natural images with $32\times 32$ pixel each. CIFAR10 contains $50,000$ training images and $10,000$ testing ones drawn from $10$ classes. CIFAR100 contains the same number of training and testing images but drawn from $100$ classes. For fair comparisons, we adopt an identical data augmentation scheme \cite{he2016deep} for all experiments, where the images are 
firstly
randomly
shifted 
with at most 
$4$ pixels along the height and width directions, and then randomly flipped horizontally. 

We implement ResNets with the IC layer according to the residual architectures shown in Fig.~\ref{fig_ResNet_archtect}(b). The network inputs are $32\times 32$ images with
the per-pixel mean subtracted and its first layer is $3\times 3$ convolutions. Then we use a stack of $6n$ layers with $3\times 3$ convolutions on the feature maps of size $\{32\times 32,16\times 16,8\times8\}$ respectively, with $2n$ layers for each feature map size. The subsampling is performed by convolutions with a stride of $2$, and the numbers of feature maps are $\{16,32,64\}$ respectively.  The network ends with a global average pooling, a $10/100$-way fully-connected layer and softmax. Thus, there are $6n+2$ stacked weighted layers in total. 

The bottleneck version of a ResNet, denoted as ResNet-B, is constructed following \cite{he2016identity}. For example, a $\left[\begin{matrix}
3\times 3,16\\
3\times 3,16
\end{matrix}\right]$ unit with two weight layers is replaced with a $\left[\begin{matrix}
1\times 1,16\\
3\times 3,16\\
1\times 1,64\\
\end{matrix}\right]$ unit with three weight layers. Thus, there are $9n+2$ stacked weighted layer in total. We pay special attentions to the last Residual Units in the entire networks of version $1$ and $2$, where we adopt an extra activation and IC layers right after its element-wise addition for version $1$, and adopt an extra IC layer for version $2$.

In this paper, we train all ResNets with depths $110(n=18)$ and $164(n=27)$ and ResNet-Bs with depths $110(n=12)$ and $164(n=18)$ by Adam for $200$ epochs with $64$ samples in a mini-batch. Following the learning-rate schedule used in archive of the Keras examples \footnote{https://github.com/keras-team/keras/blob/master/examples/cifar10\_resnet.py}, our learning rate starts from $0.001$, and is divided by $10$ at $80$, $120$ and $160$ epochs respectively. The dropout rate for all experiments is set as $0.05$. All networks are initialized following \cite{he2015delving}.

Tab. \ref{tab_ResNet} contains the testing errors of the ResNets and ResNet-Bs reformulated with the IC layer achieved at the end of training. It can be found that all three types of our residual units outperform the original ResNets and ResNet-Bs, as shown in Fig. \ref{fig_ResNet_archtect} (a), on both CIFAR10 and CIFAR100 datasets, and the residual unit with \emph{ReLU-IC-Conv2D} significantly outperforms the baseline architecture on the more challenging CIFAR100 dataset, which indicates that the representative power of ResNets and ResNet-Bs can be significantly improved by using our IC layer.

\begin{figure*}[t]
	\begin{center}
		\includegraphics[width=\linewidth]{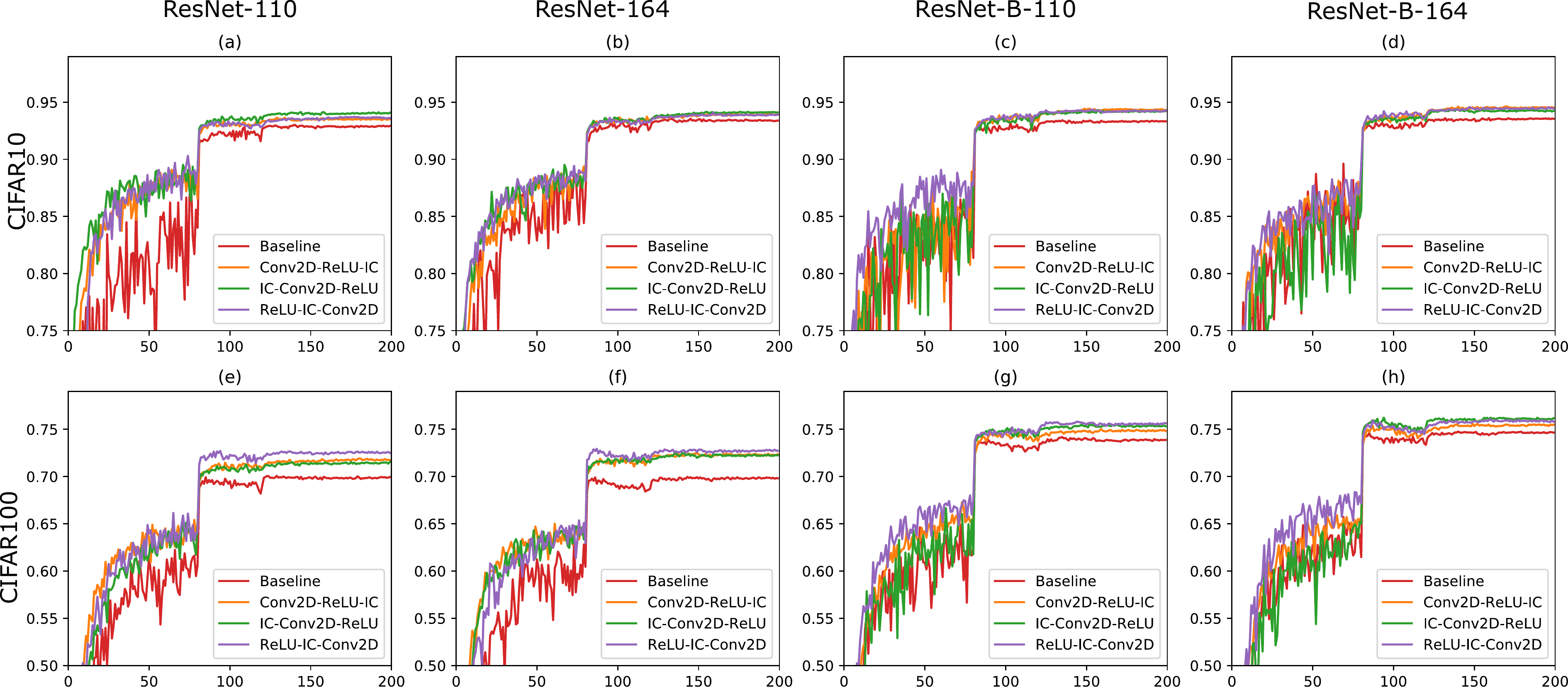}
	\end{center}
	\vspace{-4mm}
	\caption{The testing accuracy of implementing ResNet and ResNet-B with the IC layer on the CIFAR10/100 datasets with respect to the training epochs. (a) ResNet110 on CIFAR 10. (b)  ResNet164 on CIFAR 10. (c) ResNet-B 110 on CIFAR 10. (d)  ResNet-B 164 on CIFAR 10. (e) ResNet110 on CIFAR 100. (f)  ResNet164 on CIFAR 100. (g) ResNet-B 110 on CIFAR 100. (h)  ResNet-B 164 on CIFAR 100.}
	\label{fig_ResNet_res}
\end{figure*}

Fig.~\ref{fig_ResNet_res} compares the testing accuracy of our models with the baseline approach on the CIFAR10/100 datasets during the training process. As shown in Fig.~\ref{fig_ResNet_res} (a) (b) (e) (f), we find that when implemented with ResNets, all three types of our residual units achieve more stable training performance with faster convergence speed, and converges to better solutions compared with the baseline approach on both CIFAR datasets. When implemented with ResNet-Bs, our residual unit with \emph{IC-Conv2D-ReLU} has the unstable optimization process as the baseline architecture, but achieves better generalization limit. In particular, it can be found that the residual unit with \emph{ReLU-IC-Conv2D} always achieves the most stable training performance among all types of residual units implemented with the both ResNet and ResNet-B architectures on both CIFAR datasets. Thus, unless otherwise specially explained, we adopt the residual unit with \emph{ReLU-IC-Conv2D} for all rest experiments.

Note that we have included BatchNorm layers in all baseline ResNet architectures for fair comparisons. It has been demonstrated that BatchNorm can significantly smooth the optimization landscape, which induces a more predictive and stable behavior of the gradients \cite{santurkar2018does}. However, as shown in Fig.~\ref{fig_ResNet_res}, the traditional usage of BatchNorm still leads to an unstable optimization process compared with our implementation, which inspires us to question the common practice of placing BatchNorm before the activation layer. Fig.~\ref{fig_ResNet_res} verifies that our method further improves the modern networks in three ways: i) \textbf{more stable training process}, ii) \textbf{faster convergence speed}, and iii) \textbf{better convergence limit}.

\begin{figure*}[t]
	\begin{center}
		\includegraphics[width=0.8\linewidth]{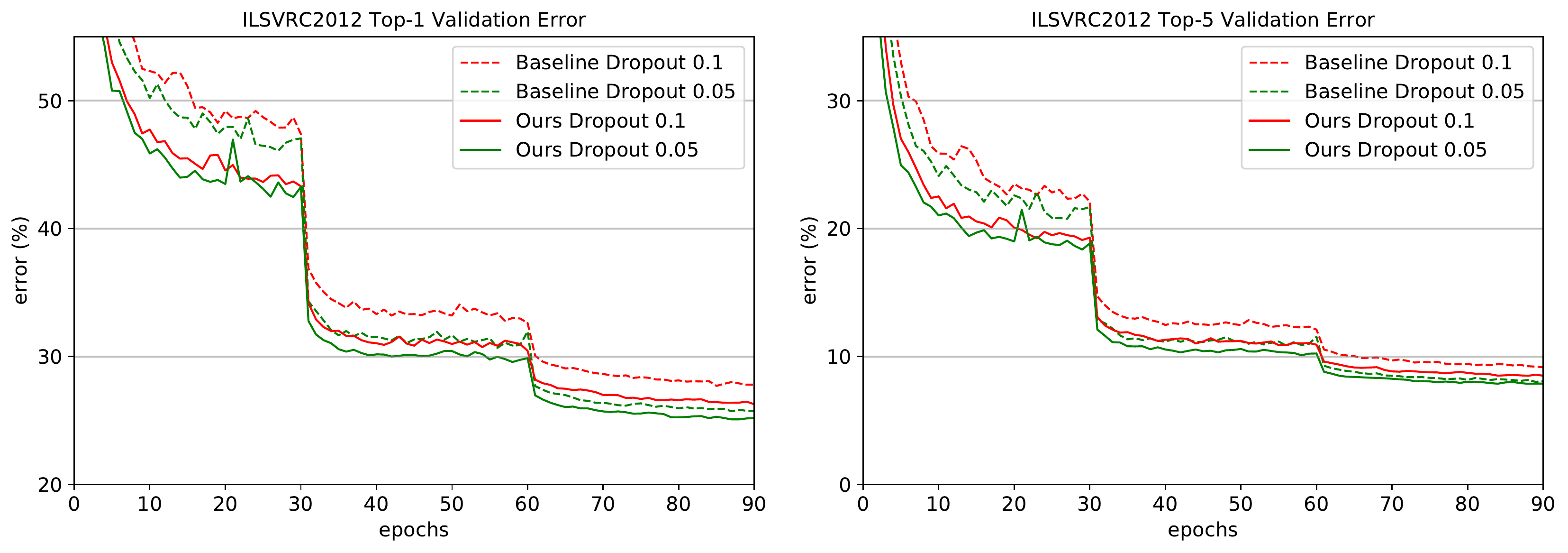}
	\end{center}
	\vspace{-4mm}
	\caption{Top-1 and Top-5 (1-crop testing) error on ImageNet validation.}
	\label{fig_ImageNet}
\end{figure*}

\subsection{ILSVRC2012}

The ILSVRC2012 classification dataset \cite{russakovsky2015imagenet} consists of $1000$ classes. We train the models on the $1.28$ million training images, and evaluate on the $50k$ validation images. Following the standard pipeline \cite{szegedy2015going}, a crop of random size of $8\%$ to $100\%$ of the original size and a random aspect ratio of $3/4$ to $4/3$ of the original aspect ratio is made, then the crop is resized to $224\times224$ with per-pixel mean subtraction for training. We use SGD with a mini-batch size of 256. The learning rate starts from $0.1$ and is decayed by $10$ every $30$ epochs, and the models are trained for $90$ epochs. When evaluating the error rates, we report both top-1 and top-5 error using the 1-crop testing - center 224x224 crop from resized image with shorter side=$256$.

In this subsection, we mainly compare our method against the implementation of \cite{li2018understanding}, which studied the problem of "\emph{why do Dropout and BatchNorm often lead to a worse performance when they are combined together?}" and found that by applying Dropout after all BatchNorm layers, most of modern networks can therefore achieve extra improvements. In \cite{li2018understanding}, the relative position of Dropout and BatchNorm layers are discussed based on the variance inconsistency caused by Dropout and BatchNorm. However, it still remain unclear whether to place Dropout and BatchNorm before or after the weight layer. \cite{li2018understanding} follows the traditional practice of placing the IC layer before the activation function.  Here, we verify that our implementation can further improve the performance compared with \cite{li2018understanding}.

In consideration of running time, we implement the ResNet-50 \cite{he2016deep} for all experiments. For our methods, we implement the IC layer according to the residual architectures shown in Fig.~\ref{fig_ResNet_archtect}(b). Specifically, we use the first version \emph{ReLU-IC-Conv2D} with a dropout rate out of $\{0.05,0.1\}$. All implementations share the same basic architecture ResNet-50 \cite{he2016deep}. The input layer is $7\times7$ convolutions, followed by a stack of "bottleneck" building blocks:
$\left[\begin{matrix}
1\times 1,64\\
3\times 3,64\\
1\times 1,256\\
\end{matrix}\right]\times3$, 
$\left[\begin{matrix}
1\times 1,128\\
3\times 3,128\\
1\times 1,512\\
\end{matrix}\right]\times6$, 
$\left[\begin{matrix}
1\times 1,256\\
3\times 3,256\\
1\times 1,1024\\
\end{matrix}\right]\times4$, 
$\left[\begin{matrix}
1\times 1,512\\
3\times 3,512\\
1\times 1,2048\\
\end{matrix}\right]\times3$. The output is simply a full-connected layer.

Fig.~\ref{fig_ImageNet} shows convergence of the validation error during training. The figure demonstrates that our method is significantly better: i) better convergence limit, ii) faster convergence speed, especially in the first $30$ epochs.  Since deeper networks are more prone to overfitting, which can be eased by the IC layer, the advantages of our method will be more significant in deeper networks such as ResNet-152. More results will be reported in future.

%

\section{Discussions}
\label{Sec_Dis}
\subsection{The Placement of BatchNorm}
The comparisons shown in Fig.~\ref{fig_ResNet_res} inspire us to take a close look at the common practice of putting a weight layer right after a ReLU activation. It can be found that
the traditional operation presents a zigzag optimization behavior and forbids the network parameters from updating in the gradient direction, which is the fastest way for the loss to achieve minimum. In this section, we would depict this behavior theoretically. 

Let $L$ be the loss of a neural net.
It is of the form 
$$L=\mathbb E_il_i,\quad l_i=D(z^i,\hat z^i),$$ 
where $D$ is some distance or divergence (e.g. $l^2$ distance, KL divergence/cross entropy). The prediction $\hat z^i$ for the $i$-th sample is calculated as
$\hat z^i=f(Wx^i)$, 
where $x^i$ is the output of an intermediate ReLU layer, $W$    
is a set of weight parameters 
applied to $x^i$, and $f$ denotes a nonlinear function approximated by the following neural network.
For convenience, we will omit the subscript $i$ for $l_i,x^i$.

Let $y = Wx$ with $W=[w_1;\ldots;w_m]\in\mathbb{R}^{m\times n}$, we calculate the sample gradient
$\partial_{w_j}l$
which is the key component of stochastic gradient descent (SGD)
$$\partial_{w_j}l={\partial l_{y_j}} x^T.$$
Since $x$ contains the outputs of a ReLU layer, each component is non-negative. This forces the updates of $w_j$ to be simultaneously positive or negative according to the sign of 
${\partial l_{y_j}}$. In other words, the incoming weights into a neuron in weight layers should only decrease or increase together for a given training sample.

However, in most situations, this direction is incompatible with the true gradient $\nabla_{w_j} L$.
Note that this is different from the sample gradient $\nabla_{w_j} l$ due to the stochastic nature of SGD. 
Indeed, when we expand (in a distributional sense)
$L=L(Wx)$ 
in $W$ around a local minimum, with the help of shift and rotation, $L$ is always of the form $\sum a_{jk}W_{jk}^2$ for some $a_{jk}>0.$ 
\footnote{One can give a more rigorous argument using a positive definite matrix.}
As a consequence, the partial derivative $\partial_{W_{jk}}L$ has the same sign as $W_{jk}$. 
Therefore, $\nabla_{w_j}L$ is parallel to $\nabla_{w_j} l$ only when
the vector $w_j\in\mathbb{R}^{1\times n}$ lies in the hyperoctants $(+,+,...,+)$ or $(-,-,...,-)$, which covers only $\frac{1}{2^{n-1}}$ portion of the region where $w_j$ could be. 
The best one could do in this situation using SGD is to follow a zigzag path as shown in \cite{lecun2012efficient}. 
Similar problems are also noticed in \cite{lecun2012efficient} when one tried to use sigmoid function as the output of a hidden layer. This old problem was resolved when sigmoid was replaced by its symmetric counterpart, namely hyperbolic tangent. While in modern applications where ReLU is more powerful and common, the issue still remains as explained above and can still be  found in our experiments. The similar results can be also observed in Fig.~\ref{fig_ImageNet}, where the IC layer stabilizes the training process and achieves better generalization performance especially in the first $30$ epochs.

\section{Conclusions and Future Works}
In this paper, we rethink the usage of BatchNorm and Dropout in the training of DNNs, and find that they should be combined together as the IC layer to transform the activation into independent components, and place this layer right before the weight layer. As shown in our theoretical analysis, the IC layer can quadratically reduce the pairwise mutual information among neurons with respect to the dropout layer parameter $p$. By reformulating ResNets with IC layers, we achieve more stable training process, faster convergence speed and better generalization performance. In future, we should consider incorporating more advanced normalization methods, such as layer normalization \cite{ba2016layer}, instance normalization \cite{DBLP:journals/corr/UlyanovVL16},  group normalization \cite{wu2018group} etc., into the IC layer, and consider more advanced statistical techniques to construct independent components. 

\section*{Acknowledgment}
\bibliography{example_paper}
\bibliographystyle{icml2019}

\end{document}